\documentclass[11pt,a4paper]{article}
\usepackage[hyperref]{acl2018}
\usepackage{times}
\usepackage{latexsym}
\usepackage{enumitem}
\usepackage{url}
\usepackage{booktabs}
\usepackage{graphicx}

\aclfinalcopy
\title{Hate Speech Detection from Code-mixed Hindi-English Tweets Using Deep Learning Models}

\author{Satyajit Kamble \\
  K J Somaiya College of Engineering \\
  Mumbai, India \\
  {\tt satyajit.k@somaiya.edu} \\\And
  Aditya Joshi \\
  CSIRO Data61\\
  Sydney, Australia\\
  {\tt aditya.joshi@csiro.au} \\}

\date{}

\begin{document}
\maketitle
\begin{abstract}
This paper reports an increment to the state-of-the-art in hate speech detection for English-Hindi code-mixed tweets. We compare three typical deep learning models using domain-specific embeddings. On experimenting with a benchmark dataset of English-Hindi code-mixed tweets, we observe that using domain-specific embeddings results in an improved representation of target groups. We also show that our models result in an improvement of about 12\% in F-score over a past work that used statistical classifiers.
\end{abstract}
\textit{This paper has been selected for publication at 15th ICON 2018, to be held in Patiala, India, in December 2018}
\section{Introduction}
Hindi is one of the official languages of India\footnote{\url{https://en.wikipedia.org/wiki/Hindi}}, spoken by more than 551 million speakers\footnote{\url{https://en.wikipedia.org/wiki/List_of_languages_by_number_of_native_speakers_in_India}}. As is typical of social media in any language, Hindi speakers on social media occasionally manifest hate towards one another. Hate speech refers to the use of hateful language, tone or prosody directed towards a person or a group of individuals, with the negative intention to provoke, intimidate, express contempt or cause harm to them. The membership to a group could be based on attributes such as race, religion, sexual orientation, ethnic origin, disability and so on.  

Hate speech detection is the automated task of detecting if a piece of text contains hate speech. Hateful messages can be used to misinform people or result in violent incidents arising due to hate, therefore, hate speech detection assumes importance. In a recent news report, the Indian Government also expressed its intention to introduce a law to deal with online hate speech\footnote{\url{https://www.thehindu.com/news/national/centre-moves-for-law-on-online-abuse/article23295440.ece}}. A tool for hate speech detection on social media in India is the need of the day.

As a country with high internet penetration and rich linguistic diversity, hate speech detection assumes an additional change in the case of Indian languages~\cite{bali2014borrowing}.
Due to the difficulties in typing tools and familiarity with the English QWERTY keyboard, using a mixture of English words and transliterated Indian language words is common amongst the Indian internet users. Referred to as code-mixing or code-switching, the phenomenon corresponds to the use of transliterated words from one or more languages along with words in the language of the script. Challenges of creating and using code-mixed datasets are well-understood~\cite{jamatia2016collecting}.

Towards this, we present an approach that uses deep learning for hate speech detection. We compare our approach with the past work by \citet{bohra2018dataset} and report a substantial improvement. The contribution of our work is:
\begin{enumerate}
\item We compare our deep learning-based approach with a statistical approach, and evaluate it on the same dataset as the statistical approach. We observe an improvement in the performance.
\item Instead of using pre-trained word embeddings, we train word embeddings on a large corpus of relevant code-mixed data. We demonstrate that this results in improved similarity values.
\end{enumerate}
The rest of the paper is organised as follows. We describe related work in Section~\ref{sec:relwork}. The architecture is in Section~\ref{sec:archi} while the experiment setup is in Section~\ref{sec:expsetup}. We present our results in Section~\ref{sec:res}, and analyse the errors in Section~\ref{sec:erroranal}.

\section{Related Work}
\label{sec:relwork}
Approaches for hate speech detection have been reported~\cite{schmidt2017survey,warner2012detecting}. Code-mixed datasets for Indian languages have been explored for several NLP tasks such as part-of-speech tagging~\cite{jamatia2015part}, language identification~\cite{das2014identifying} and so on. Also, work concerning with hate speech in English language exists~\cite{waseem2016hateful,djuric2015hate,davidson2017automated,nobata2016abusive}. In a way, code-mixed datasets represent a majority of datasets from India, on the social media. \citet{bohra2018dataset} introduces a dataset of Hindi-English code-mixed tweets, and reports results on a statistical approach that use hand-engineered features. We download tweets from their dataset and compare with their results. Another work by \citet{mathur2018detecting} uses deep learning for hate speech detection. Our work differs from theirs in two ways: (a) We experiment with a different dataset, and compare performance on that dataset with the past work that reports results on the dataset, (b) We use domain-specific word embeddings that we show to be better indicative of semantics in the hate speech context.
\begin{figure}
\centering
\includegraphics[width=0.9\linewidth]{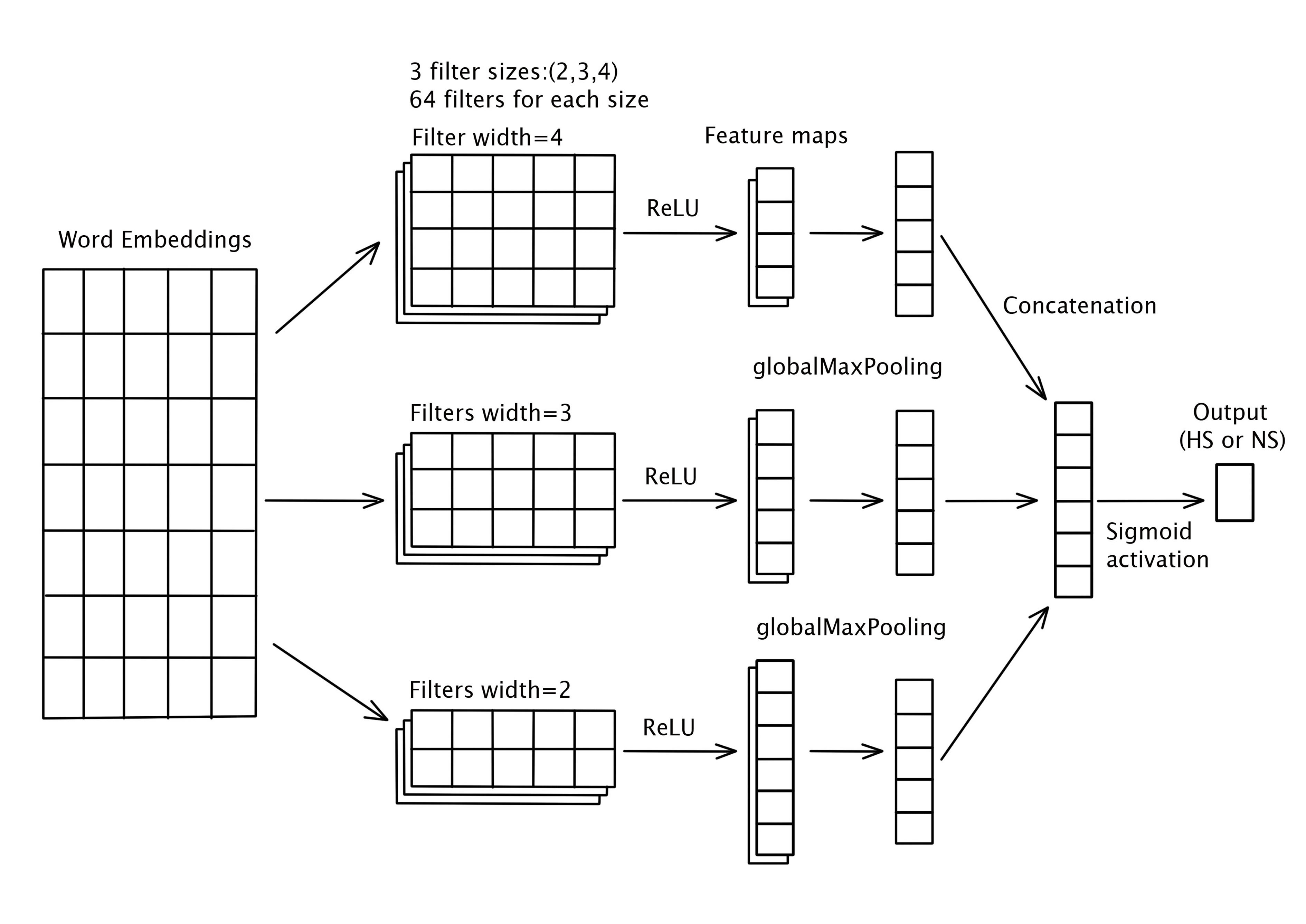}
\caption{CNN model for hate speech detection.}
\label{fig:cnn}
\end{figure}
Our approach of using domain-specific embeddings is motivated by ~\citet{tkachenko2018searching}. They train two sets of word embeddings: one from a Wikipedia corpus and another from an Amazon review corpus. For sentiment-related tasks (such as sentiment classification), embeddings on the Amazon review corpus result in a higher performance as compared to those from the Wikipedia corpus. On the other hand, for topic-related tasks (such as topic classification), embeddings trained using the Wikipedia corpus outdo those from the Amazon review corpus. 

\section{Architecture}
\label{sec:archi}
We propose three deep learning models for hate speech detection. These models are shown in Figures~\ref{fig:cnn}, ~\ref{fig:lstm} and ~\ref{fig:bilstm} respectively. In the forthcoming sections, we describe each of the models.
\subsection{CNN-1D} 
Figure~\ref{fig:cnn} shows the CNN-1D model. It is fed in with domain-specific embeddings corresponding to sentences in the training data. The filters(3 filter sizes) with the specifications listed, convolve over the embeddings and produce the feature maps. Following this, we use a layer of globalMaxPooling having a dropout probability of 0.5. Then, the results are concatenated to form a single feature vector. Here, we apply the sigmoid activation to produce our final results.

\subsection{LSTM}
\begin{figure}
\centering
\includegraphics[width=\linewidth]{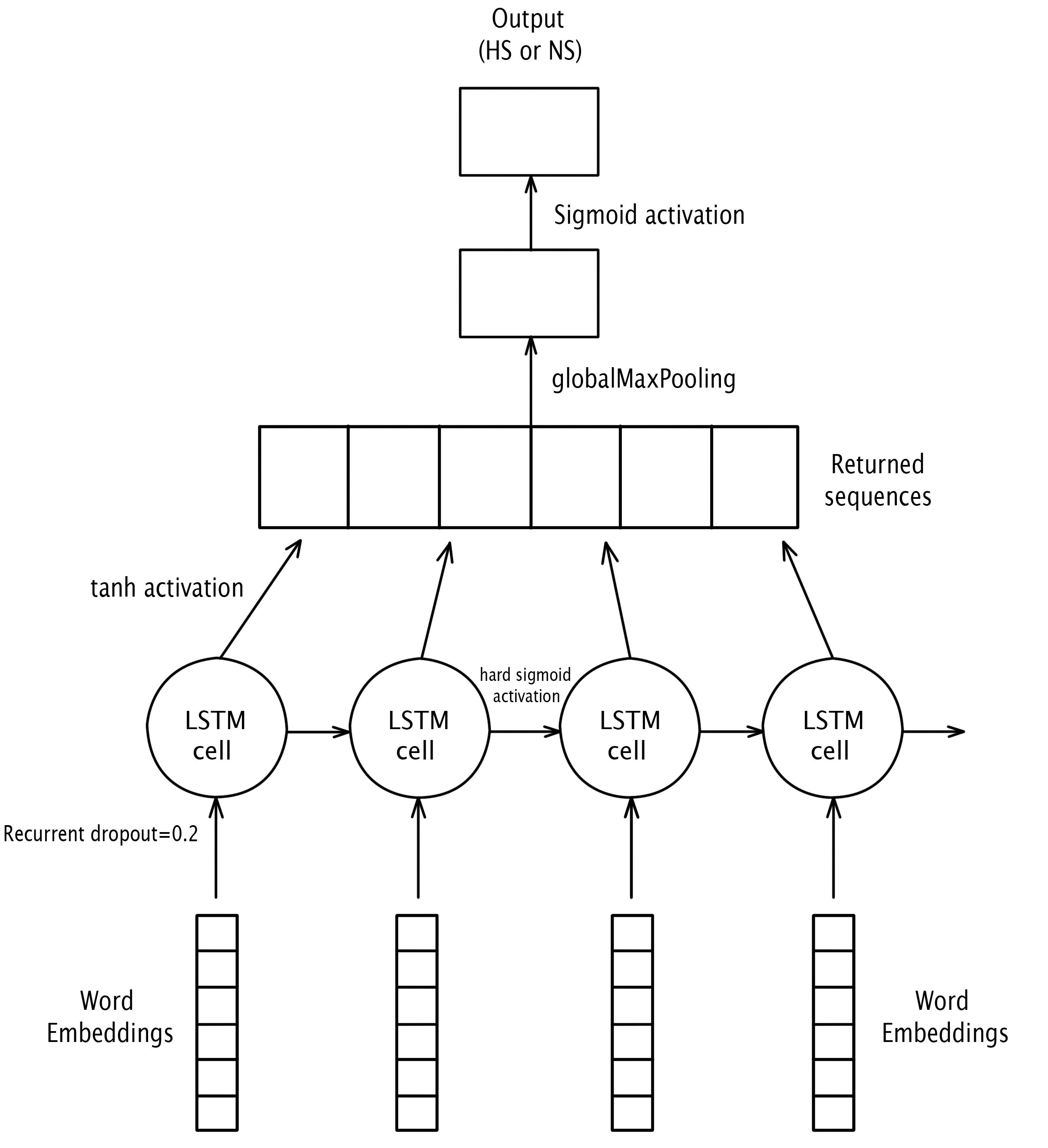}
 \caption{LSTM model for hate speech detection.}
\label{fig:lstm}
\end{figure}

Figure~\ref{fig:lstm} shows the LSTM model. Owing to the sequential nature of the code-mixed data, we make use of the LSTM model to compare our results. The results of the input embeddings, on passing through the LSTM layer, are made to accumulate at each proceeding timestep. The model is tuned to return the sequences of each of these timesteps. Next, the compiled sequences are given as an input to the globalMaxPooling layer. Lastly, the resulting output from the pooling layer is passed through the sigmoid activation function to give a final prediction.
\begin{figure}
\centering
\includegraphics[width=\linewidth]{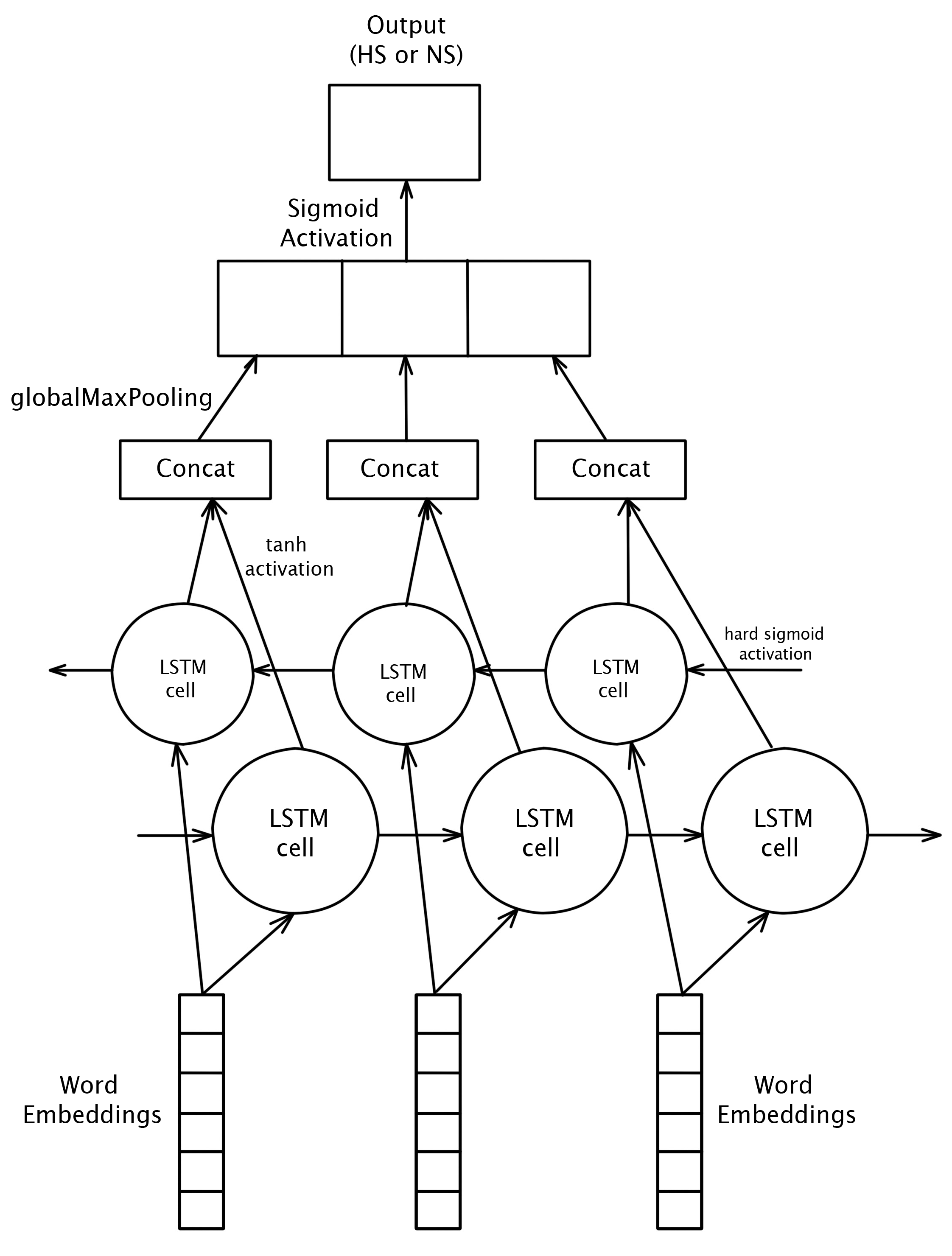}
\caption{BiLSTM model for hate speech detection}
\label{fig:bilstm}
\end{figure}
\subsection{BiLSTM} 
Figure~\ref{fig:bilstm} shows the BiLSTM model. Taking into consideration that the temporal dynamics can be better captured when a piece of text is analysed from both the directions, we make use of the BiLSTM to further compare our results. Here, instead of retrieving the sequences from a single direction, we do it for both the directions and concatenate the results. The vector now produced, goes through the globalMaxpooling layer. Finally, the result produced, is passed through the sigmoid activation to generate the final output.
\subsection{Creation of Domain-Specific Word Embeddings}
Using the Twitter API, we search for tweets containing Hindi cuss words and names of minority groups in their transliterated form. This is motivated by the definition of hate speech: hateful language that is used towards minority groups. We download a dataset of 255,309 tweets. Statistics of the dataset are in Table~\ref{tab:stats}. Tweets collected were used only to train word embeddings. The dataset by \citet{bohra2018dataset} is used for evaluation of the 3 deep learning models. 
 \begin{table}[]
 \centering
\begin{tabular}{lp{1.65cm}p{1.65cm}}\toprule
Dataset Charecteristics  & Size \\ \midrule
Number of Tweets & 255,309\\
Number of Timelines Extracted & 7232 \\
Number of Retweets & 76,645 \\
Total Number of Words & 4,975,642 \\
Size of Vocabulary & 168,638 \\
\% Hindi Words per Tweet & 18.63\%
\\\bottomrule
\end{tabular}
\caption{Dataset Statistics of the Domain-Specific Word Embeddings}
\label{tab:stats}
\end{table}

We use the gensim (\url{https://radimrehurek.com/gensim/models/word2vec.html}) library to train word embeddings from this dataset, and use these domain-specific embeddings to initialize our deep learning models. We also utilize the Google Translate API to measure the average Hindi proportion of all the collected tweets. Using the API, we calculate the number of Hindi words in a tweet and calculate it's percentage with respect to the total number of words in the tweet. This is done for all tweets and an average is computed. \textit{We commit to make our domain-specific word embeddings available for download at: \url{https://github.com/satyaSK/Hate-Speech-Detection}}.

\section{Experiment Setup}
\label{sec:expsetup}
We download the dataset by \citet{bohra2018dataset} using the Twitter API. Due to typical issues such as timeline restrictions, we obtain \textit{3849} tweets, of which \textit{1436} are labelled as hateful. We report 10-fold cross-validation performance on this dataset. We compare our models with a baseline re-implementation as given in \citet{bohra2018dataset}. We implement feature extraction and use classification algorithms as described in their paper. 

For the deep learning models, we use Keras, a neural network API (\url{https://keras.io/}). We experimentally determine the values of the parameters. For the CNN-1D model, we use the following hyperparameters: 
\begin{enumerate} \itemsep\setlength{0em}
\item Embedding dimension = 300
\item Number of filters of each filter size = 64, Batch size = 64, Epochs = 5, Dropout = 0.5
\item Pooling layer : Global max pooling 
\item Filter sizes being 2,3 and 4 for the 3 CNNs in parallel.
\item Loss function : Binary cross-entropy loss
\item ReLU activation to obtain feature maps  
\item Optimization algorithm : Adam
\end{enumerate}
For LSTM and BiLSTM, we use the following configuration:
\begin{enumerate} \itemsep\setlength{0em}
\item Number of LSTM units = 100, Recurrent dropout = 0.2
\item Loss function : Binary cross-entropy loss
\item Recurrent Activation : Hard sigmoid
\item Activation : tanh
\end{enumerate}
We report Precision, Recall, F-score and accuracy values using methods in scikitlearn\cite{pedregosa2011scikit}.
\section{Results}
\label{sec:res}
\subsection{Qualitative Evaluation of domain-specific word embeddings}
Table~\ref{tab:embeddingsim} shows cosine similarity between `women' and words of three minority groups: religious, caste and sexual. We have not mentioned the specific names of the corresponding groups due to their controversial nature.  \textit{We wish to highlight that the word `women' is used as a reference word solely because women might be a target of hate speech on social media}. Each row in the table is computed using the cosine similarity between the word `women' and representative words of the specific minority group. The similarity between a pair of related social groups is consistently higher in the case of domain-specific embeddings as compared to general embeddings. For example, in case of sexual minority (which we consider as `\textit{transgender}'), the similarity in the case of domain-specific embeddings is 0.726 while that in case of general embeddings is 0.348. This implies that domain-specific embeddings are able to capture the societal relationships and correlations between minority groups more accurately. An additional point to note is that, swear words in Hindi may not be present in pre-trained Google news embeddings. Specifically, we observe that 18 swear words in Hindi that were used to download the dataset, and were used to train domain-specific embeddings are not present in the Google news embeddings at all. 

Therefore, higher similarity between groups that are targets of hate speech and higher coverage in terms of words that indicate expressions of hate, highlight the importance of using domain-specific embeddings.
\subsection{Quantitative evaluation of hate speech detection}
 \citet{bohra2018dataset} train their classifiers using SVM and Random Forest algorithm, but only report accuracy. For a better comparison, we re-implement their features and obtain Precision, Recall and F-score values as well. The reported values and our values are compared with the deep learning models in Table~\ref{tab:res}. It must be noted that the accuracy values as reported and as obtained from re-implementation are close - indicating that the precision and recall are also likely to be comparable.
 \begin{table}[]
 \centering
\begin{tabular}{lp{1.65cm}p{1.65cm}}\toprule
Minority Group   & Domain-specific & General \\ \midrule
Religious Minority & 0.637 & 0.224 \\
Caste Minority & 0.615 & 0.204 \\
Sexual Minority & 0.726 & 0.348 \\\bottomrule
\end{tabular}
\caption{Cosine Similarity of `women' with words representing three minority groups.}
\label{tab:embeddingsim}
\end{table}
 We observe that using CNN-1D results in the highest performance with a F-score of 80.85\% and an accuracy of 82.62\%. This improvement in F-score is about 12\% higher than the statistical baseline that we compare against. The improvement is in both precision and recall. An example of a correctly classified instance of hate speech by the CNN-1D model is `\textit{@.. inke 6month ke works dekh lijiy nafrat ho jayegi aapko inse anandpal ke liye julus aur julus me public ko khule aam patthro ki barish karna dhamkana public ke sir fodna hate all of u}' which is translated to `\textit{@.. look at the 6 month works of these people, you will start to hate them. A group of people rallying for Anandpal, has been stone-throwing and threatening the public. hate all of you.}`. Among the deep learning models, we observe that CNN-1D results in the highest precision while BiLSTM gives the best recall by a difference of approximately 0.40\% as compared to the CNN-1D. For example, this tweet `\textit{@.. he is right x y may gundo ka palka kutha hai jo koi karwai nai kartha gundo par u.p no1 state in muders rape}` \textit{(@.. he is right, x is a dog pet by the mafias of y, and so, he does not call for the investigation of the crimes they committed. u.p is number 1 state in rape and murders)} has been correctly classified as hate speech by the CNN-1D model while the LSTM and BiLSTM models incorrectly classify the tweet as non-hate speech.(x and y are anonymised names of a politician and a state respectively). In general, these results show that our deep learning models outperform the statistical approach.
\begin{table}[]
\centering
\begin{tabular}{p{2cm}lllp{1cm}}
\toprule
& \textbf{P (\%)}  & \textbf{R (\%)}          & \textbf{F (\%)}  & \textbf{A (\%)} \\ \midrule
%\multicolumn{5}{c}{\textbf{Baseline}}\\ \midrule
%Reported value for SVM by \cite{bohra2018dataset}                 & \multicolumn{3}{c}{\textit{-not reported in the original paper-}} & 71.7     \\
%Reported value for Random Forest by \cite{bohra2018dataset} & \multicolumn{3}{c}{\textit{-not reported in the original paper-}} & 66.7     \\ \hline
\cite{bohra2018dataset} (SVM)           & 74.94               & 63.15           &   68.54               & 71.03 (71.7*)    \\
\cite{bohra2018dataset} (Random Forest) & 62.43               & 58.88           &   60.60               & 65.78 (66.7*)    \\ \midrule
%\multicolumn{5}{c}{\textbf{Our Deep Learning-based Models}} \\ \midrule
CNN-1D                                                           & \textbf{83.34}               & 78.51           &  \textbf{80.85}      & \textbf{82.62}    \\
LSTM                                                             & 81.11               & 75.80           &  78.36     & 80.21    \\
BiLSTM                                                           & 82.04               & \textbf{78.90}           &  80.43 & 81.48   \\ \bottomrule
\end{tabular}
\caption{Comparison of Statistical Approach with Our Deep Learning-based Approach for Hate Speech Detection; * indicates reported values in the baseline paper; P: Precision, R: Recall, F: F-score, A: Accuracy.}
\label{tab:res}
\end{table}
\section{Error Analysis}
\label{sec:erroranal}
To understand the shortcomings of our models, we analyse and elucidate the errors made by our best-performing approach, which motivate future directions of experimentation. Some of these errors include:
\begin{itemize} \itemsep \setlength{0em}
\item \textbf{Code-switched tweets in Hindi}: These are tweets written, following the grammatical structure of Hindi with a few English words. Many mis-classified examples include such tweets. An example is `\textit{@.. @.. @.. @..  aur tum jahan hoti ho wahan balatkar badh jata hai baba bhi rape karne lagte hin (sic)}'. This tweet is translated as `\textit{@.. @.. @.. @.. and rape cases start to increase wherever you go, baba also starts to rape'}. This has been identified to be a recurring error which occurs due to the code-mixed nature of the data at hand, where the text piece contains an imbalance between tokens from the Hindi and English scripts. 

\item \textbf{Series of swear words}: Some mis-classified instances are a string of swear words with a few function words between them. We skip an example here, on purpose, due to the obscene nature of these tweets. These errors may be because the model does not solely rely on the presence of swear words. Other context may be necessary to detect hate speech. This shows that the presence of explicit hate keywords or swear words is not the only determining factor for deciding whether a piece of text is hate speech or not, which points towards the necessity of capturing the underlying semantics and sense of the text in discussion. 
\item \textbf{Possibly incorrect labels}: Some tweets contain swear words but are not hateful towards any group as such. So, even though our models predict them as non-hate-speech, the instance is marked as mis-classified. For example, a hateful tweet calls someone the child of a rape victim but the gold label is negative. On the other hand, `\textit{x ke samay me isase double rape hote the lekin us samay y bolti thi hai na (In times of x, the number of rapes were double as this, but y would always call it out, isn't it?!)} has the gold label as positive. (x and y are anonymised names of politicians).

\end{itemize}
\section{Conclusion \& Future Work}
\label{sec:concl}
This paper explored hate speech detection in Hindi-English code-mixed tweets. We used three typical deep-learning models for detecting hate speech and empirically demonstrated their effectiveness. In contrast to statistical methods, our models were able to better capture the semantics of hate speech along with their context. We additionally demonstrated the efficacy of domain-specific word embeddings in adding intrinsic value to the code-mixed landscape.

Our work uses a benchmark dataset, and shows how deep learning models improve best-known work using statistical classifiers. In that, we make a small contribution to hate speech detection for Hindi-English code-mixed tweets. Novel deep learning techniques capable of assimilating textual cues more accurately, can be used to improve upon our work. Other nuances of hate speech in terms of sarcasm or misinformation can also be incorporated in future work.
\bibliography{emnlp2018}
\bibliographystyle{acl_natbib_nourl}
\end{document}